\documentclass[10pt,twocolumn,letterpaper]{article}

\usepackage{cvpr}
\usepackage{times}
\usepackage{epsfig}
\usepackage{graphicx}
\usepackage{amsmath}
\usepackage{amssymb}
\usepackage{authblk}

\usepackage{array}
\newcolumntype{L}[1]{>{\raggedright\let\newline\\\arraybackslash\hspace{0pt}}m{#1}}
\newcolumntype{C}[1]{>{\centering\let\newline\\\arraybackslash\hspace{0pt}}m{#1}}
\newcolumntype{R}[1]{>{\raggedleft\let\newline\\\arraybackslash\hspace{0pt}}m{#1}}

\cvprfinalcopy % *** Uncomment this line for the final submission

\def\cvprPaperID{4013} % *** Enter the CVPR Paper ID here

\ifcvprfinal\fi
\begin{document}

%%%%%%%%% TITLE
\title{AANet: Attribute Attention Network for Person Re-Identifications}

\author[1,2]{Chiat-Pin Tay}
\author[2]{Sharmili Roy}
\author[1,2]{Kim-Hui Yap}
%\author{Chiat Pin Tay\\
\affil[1]{School of Electrical and Electronic Engineering, Nanyang Technological University, Singapore}
\affil[2]{Rapid-Rich Object Search Lab, Nanyang Technological University, Singapore}
%School of Electrical and Electronic Engingeering, Nanyang %Technological University, Singapore\\
%Institution1 address\\
%{\tt\small firstauthor@i1.org}
% For a paper whose authors are all at the same institution,
% omit the following lines up until the closing ``}''.
% Additional authors and addresses can be added with ``\and'',
% just like the second author.
% To save space, use either the email address or home page, not both
%%\and
%%Second Author\\
%%Institution2\\
%%First line of institution2 address\\
%%{\tt\small secondauthor@i2.org}
%}

\maketitle
\thispagestyle{empty}

%\begin{multicols}{2}

%%%%%%%%% ABSTRACT
\begin{abstract}

   This paper proposes Attribute Attention Network (AANet), a new architecture that integrates person attributes and attribute attention maps into a classification framework to solve the person re-identification (re-ID) problem. Many person re-ID models typically employ semantic cues such as body parts or human pose to improve the re-ID performance. Attribute information, however, is often not utilized. The proposed AANet leverages on a baseline model that uses body parts and integrates the key attribute information in an unified learning framework. The AANet consists of a global person ID task, a part detection task and a crucial attribute detection task. By estimating the class responses of individual attributes and combining them to form the attribute attention map (AAM), a very strong discriminatory representation is constructed. The proposed AANet outperforms the best state-of-the-art method \cite{Sun_2018_ECCV} using ResNet-50 by 3.36\% in mAP and 3.12\% in Rank-1 accuracy on DukeMTMC-reID dataset. On Market1501 dataset, AANet achieves 92.38\% mAP and 95.10\% Rank-1 accuracy with re-ranking,  outperforming~\cite{kalayeh2018human}, another state of the art method using ResNet-152, by 1.42\% in mAP and 0.47\% in Rank-1 accuracy. In addition, AANet can perform person attribute prediction (e.g., gender, hair length, clothing length etc.), and localize the attributes in the query image.

\end{abstract}

%%%%%%%%% BODY TEXT
\section{Introduction}

%Conventionally, person re-ID aims to retrieve images of a person taken from an array of cameras, or from same camera but from different occasions, by using a query image. Usually this prediction task is limited to a short time frame and within a small geographical area. This seemingly simple recognition task to human is a challenge to the machine, as factors like person occlusions, pose variation, ambient light changes, low image resolution, etc. degrade image quality and reduce effective number of usable pixels for image processing. In recent years, deep learning has overcome many limitations, and improves the retrieval performance significantly. Still, there are many rooms for improvement.

Given a query image, person re-ID aims to retrieve images of a queried person from a collection of network-camera images. The retrieval is typically attempted from a collection of images taken within a short time interval with respect to the queried image. This supports the underlying assumption that the query person's appearance and clothing attributes remain unchanged across the query and the collection images. Person re-ID is a challenging problem due to many factors such as partial/total occlusion of the subject, pose variation, ambient light changes, low image resolution, etc. Recent deep learning based re-ID solutions have demonstrated good retrieval performance. 

%\par One important person re-ID application is in the camera surveillance domain. Tens of thousands of cameras are being installed in cities all over the world, and these cameras are expected to not just record videos, but also perform identification of person attribute. Some examples of useful person attribute information are gender, hair, hat, clothing colors and types and carried bags. This prediction ability, however, is lacking in most person re-ID network.

\begin{figure}
\centering
\includegraphics[width=\linewidth]{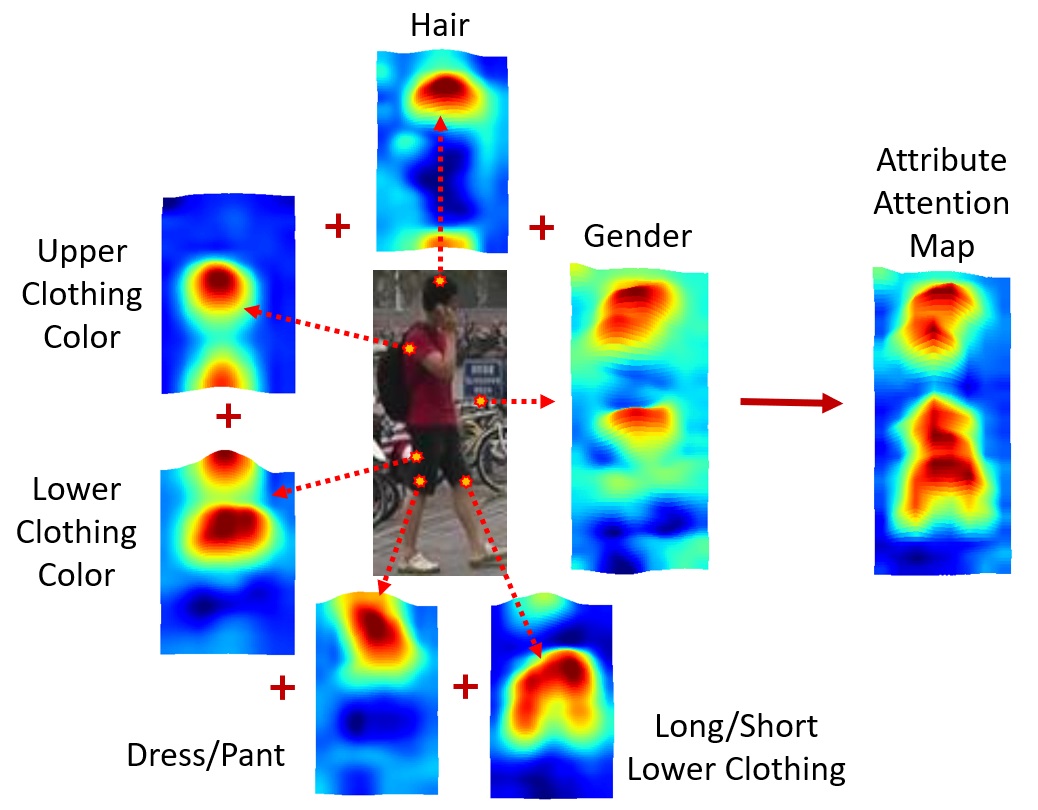}
\caption{Class-aware heat maps are extracted and combined to form a discriminatory Attribute Attention Map (AAM) at the image level. The six heat maps shown here correspond to the six attributes such as hair, upper clothing color, lower clothing color etc. Best viewed in color.}
\label{fig:MergedAttribute}
\end{figure}

\begin{figure*}
\centering
\includegraphics[width=\linewidth]{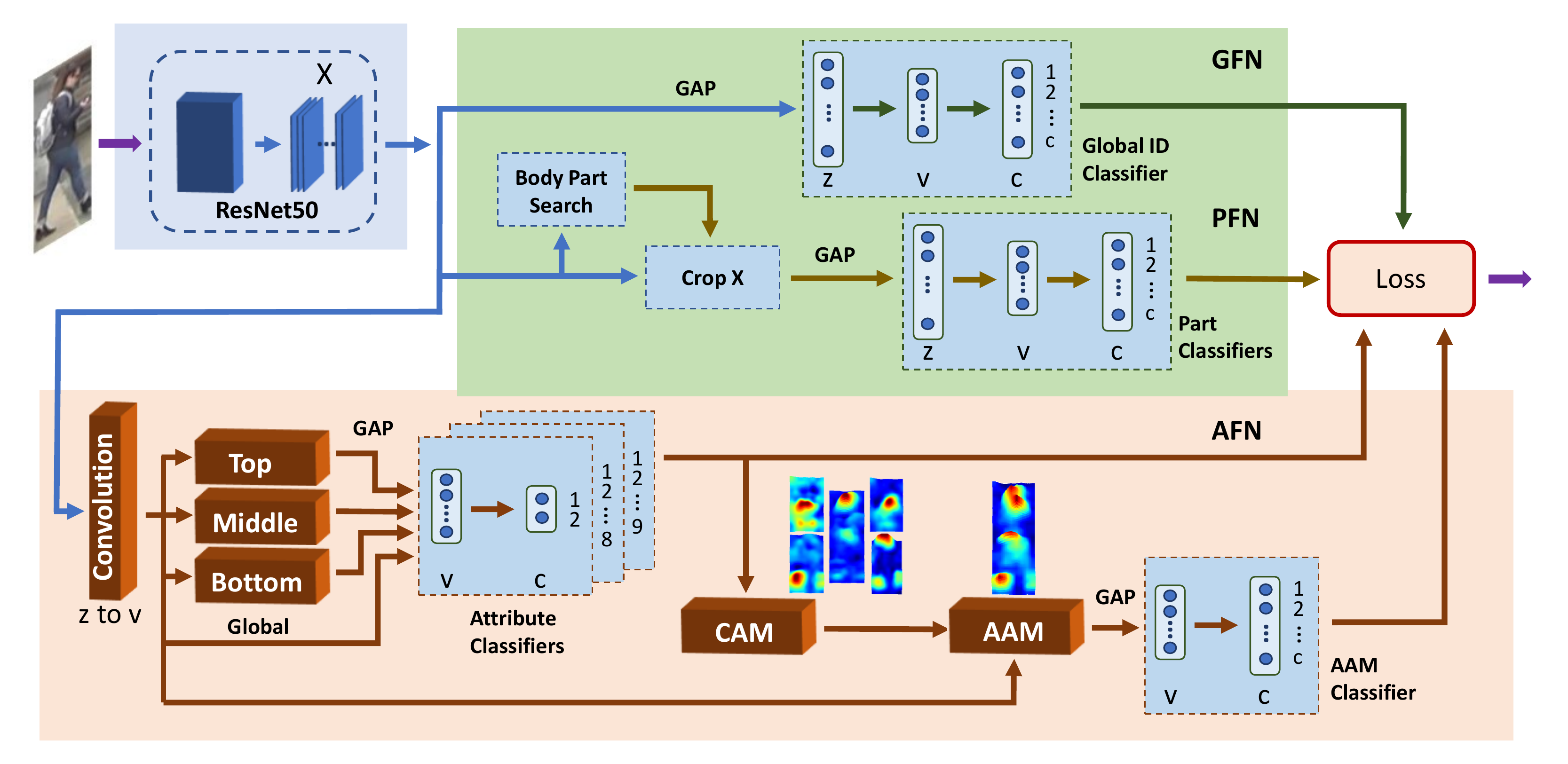}
\caption{Overview of AANet. The backbone network, which is based on the ResNet-50 architecture, outputs the feature map \(X\). The feature map \(X\) is forwarded to three tasks, namely the Global Feature Network(GFN), Part Feature Network (PFN) and Attribute Feature Network (AFN). The output of these three tasks are combined using homoscedastic uncertainty learning to predict the person identification. Best viewed in color.}
\label{fig:AANet}
\end{figure*}

%\par In this work, we aim for good person re-ID performance with person attribute prediction capability. To achieve our goals, we look to multitask network, with a person identification (ID) main task, and two relevant  attribute and body part detection sub tasks. The main task performs global feature learning. The two sub tasks, however, perform localized and different feature learning. By sharing their learning experience with the main task, we have a very discriminatory representations at the output of the network. This is shown in Figure \ref{fig:AANet}. AANet consists of two main networks. The first main network performs global identity classification, and body part detection and classification. This two tasks are managed by the Global Feature Network (GFN) and the Part Feature Network (PFN). The second main network is the Attribute Feature Network (AFN), which also consists of two tasks, namely the person attribute classification and AAM classification. All four tasks utilizes Softmax classification outputs, using image labels and person attribute labels as ground truth for learning. To ensure optimal performance, we use uncertainty learning to obtain the weighting for each task. The final task loss is then the summation of the scaled task losses.

\par The approaches used to solve the person re-ID problem can be broadly divided into two categories. The first category comprises of metric learning methods that attempt to learn an embedding space which brings images belonging to a unique person close together and those belonging to different persons far away. Various approaches such as triplet and quadruplet losses have been employed to learn such embedding spaces \cite{DBLP:journals/corr/HermansBL17, chen2017beyond}.

\par The second category of methods poses the re-ID problem in a classification set-up. Such methods learn by using Softmax normalization and computing cross-entropy loss, based on person identity as ground truth, for back-propagation. Research has shown that by integrating semantic information such as body parts, human pose etc, the classification and recognition accuracy can be significantly improved \cite{DBLP:conf/iccv/SuLZX0T17, DBLP:conf/mm/WeiZY0T17, DBLP:journals/corr/YaoZZLT17}. Person attributes, such as clothing color, hair, presence/absence of backpack, etc. are, however, not used by the current state-of-the-art re-ID methods. Since, a typical re-ID model assumes that the physical appearance of the person of interest would not significantly change between query image and the search images, physical appearance becomes a key information that can be mined to achieve higher re-ID performances. Such information is not utilized by current research and the state of the art re-ID methods. 
\par In this work, we propose to utilize the person attribute information into the classification framework. The resulting framework, called the Attribute Attention Network (AANet), brings together identity classification, body part detection and person attribute into an unified framework that jointly learns a highly discriminatory feature space. The resulting network outperforms existing state of the art methods in multiple benchmark datasets.  

\par Figure \ref{fig:AANet} gives an overview of the proposed architecture. The proposed framework consists of three sub-networks. The first network, called the Global Feature Network (GFN), performs global identity (ID) classification based on the input query image. The second network, called the Part Feature Network (PFN), focuses on body part detection. The third network is the Attribute Feature Network (AFN), which extracts class-aware regions from the person’s attributes to generate Attribute Attention Map (AAM). This is shown in Figure \ref{fig:MergedAttribute}. The three networks perform classification using person ID and attribute labels We use homoscedastic uncertainty learning to optimize the weights of the three sub-tasks for final loss calculations.

\par Since AANet performs person attribute classification as part of network learning, it also output attribute predictions for each query and gallery images. This enables attribute matching of the gallery images, with or without retrieval by query image.

\par Our key contributions can be summarized as follows:
\begin{enumerate}
    \item We provide a new network architecture that integrates attribute features with identity and body part classification in a unified learning framework.

    \item We outperform the existing best state-of-the-art re-ID method on multiple benchmark datasets and propose the new state of the art solution for person re-ID.
\end{enumerate}

The rest of the paper is organized as follows. Section~\ref{sec:relatedWorks} provides an overview of the related works. In section~\ref{sec:aanet} we describe the proposed AANet framework. Experimental results are provided in section~\ref{sec:results} and \ref{sec:results-attr}. The paper is concluded in section~\ref{sec:conclu}. 

\begin{figure}
\centering
\includegraphics[width=\linewidth]{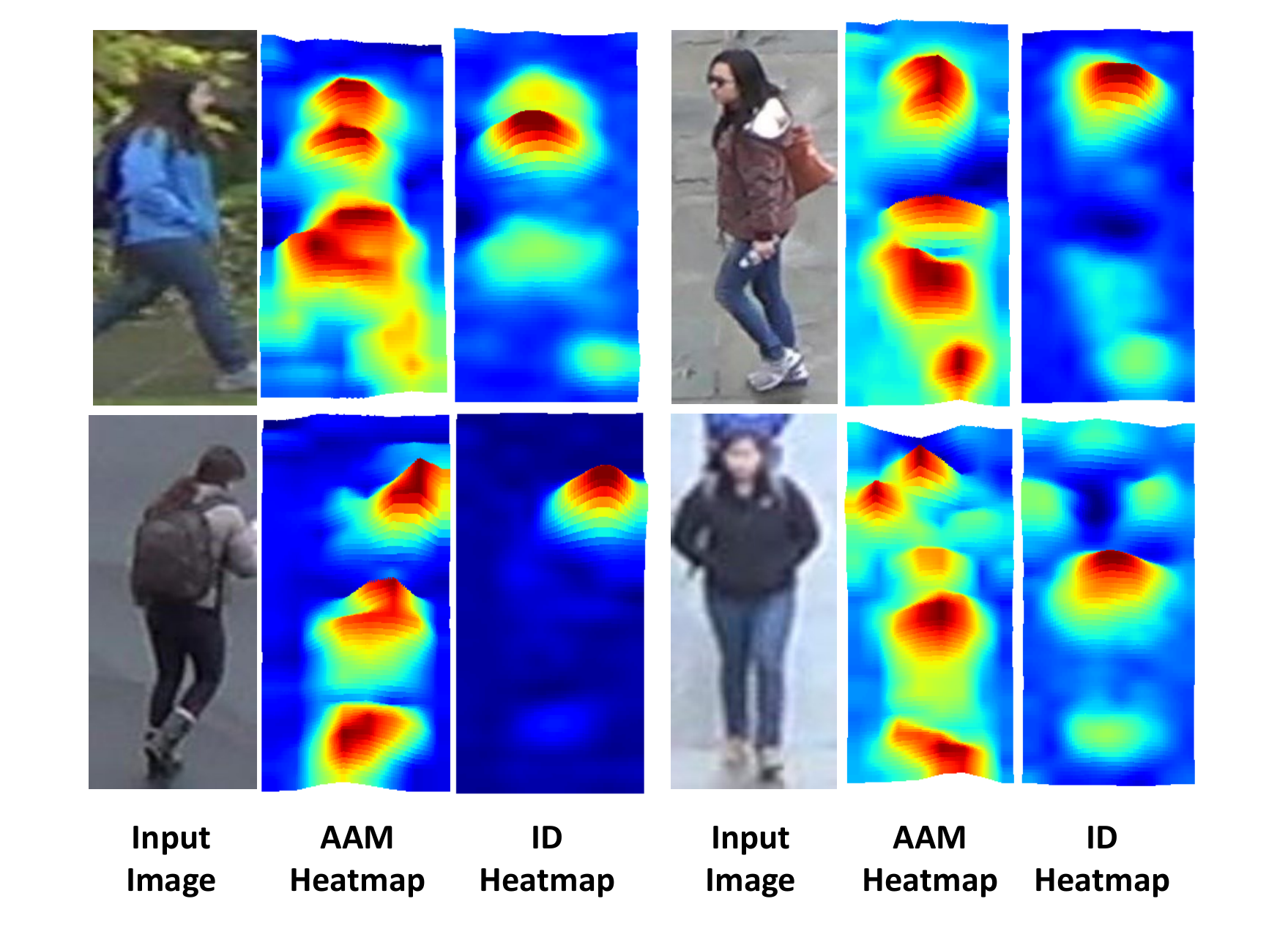}
\caption{Comparison between proposed AAM and class activation ID heatmap generated in GFN. The AAM captures person attributes, therefore the activated areas lie mostly within the pedestrian body. The ID heatmap is influenced by the training dataset, is less dense and may include background as part of its feature map. Best viewed in color.}
\label{fig:AAM_Heatmap}
\end{figure}

\section{Related works}
\label{sec:relatedWorks}
%\subsection{Hand-crafted methods and Deep Learning}
%Person re-ID has always been a challenging task for computer vision community. In the past, hand-crafted methods were used. For example, Ma \etal \cite{DBLP:conf/bmvc/MaSJ12} used Gabor filter and Covariance descriptors to compute the representations. The performance was way below the current methods. With the emergence of deep learning (DL), Wu \etal \cite{DBLP:conf/wacv/WuCLWYZ16} integrated hand-crafted algorithms into CNN to outperform prior works. Nowadays most researchers move away from the traditional methods totally, and fully embrace DL. 

%In recent years many approaches have been explored to solve the re-ID problem using traditional computer vision and deep learning based methods. In this section, we provide an overview of the recent deep learning based methods that achieve close to state-of-the-art performance.

In recent years, deep learning was used to solve various challenging computer vision tasks 
\cite{lecun2015deep, he2017mask, DBLP:journals/corr/HermansBL17, Ding_2018_CVPR, ding2019semantic}. In this section, we provide an overview of the recent re-ID deep learning based methods that achieve close to state-of-the-art performance.

Deep learning based re-ID solutions are often posed as an identity classification problem. Authors in \cite{DBLP:conf/cvpr/XiaoLOW16} used multi-domain datasets to achieve high re-ID performance in a classification set-up. Wang \etal \cite{DBLP:conf/cvpr/WangZLZZ16} proposed to formulate the re-ID problem as a joint learning framework that learns feature representations using not only the query image but also query and gallery image pairs. Many solutions have used additional semantic cues such as human pose or body parts to further improve the classification performance. Su \etal \cite{DBLP:conf/iccv/SuLZX0T17} proposed a Pose-driven Deep Convolutional (PDC) model to learn improved feature extraction and matching models from end-to-end. Wei \etal \cite{DBLP:conf/mm/WeiZY0T17} also adopted the human pose estimation, or key point detection approach, in his Global-Local-Alignment Descriptor (GLAD) algorithm. The local body parts are detected and learned together with the global image by the four-stream CNN model, which yields a discriminatory and robust representation. Yao \etal \cite{DBLP:journals/corr/YaoZZLT17} proposed the Part Loss Networks (PL-Net) to automatically detect human parts and cross train them with the main identity task. Zhao \etal \cite{DBLP:conf/iccv/ZhaoLZW17} follows the concept of attention model and uses a part map detector to extract multiple body regions in order to compute their corresponding representations. The model is learned through triplet loss function. Sun \etal \cite{Sun_2018_ECCV} proposed a strong Part-based Convolutional Baseline method, with Refined Part Pooling method to re-align parts for high accuracy performance. Kalayeh \etal \cite{kalayeh2018human} used multiple datasets, deep backbone architecture, large training images, and human semantic parsing to achieve good accuracy results. Similarly, Jon \etal \cite{almazan2018re} proposed using deep backbone architecture and large input image, but with classification as first pass learning, followed by metric learning for accuracy fine-tuning.

Integrating semantic information such as body parts and pose estimation have shown significant improvement in re-ID performance. Since a person's attributes do not change significantly between the query image and the gallery images, we believe that attributes form a key information that can significantly impact person re-ID performance. This, however, has not been utilized in the current re-ID methods. In view of this, we propose to integrate physical attributes to the identity classification framework.

%Zhou \etal \cite{zhou2016learning} proposed the Class-Activation Mapping (CAM) technique by using Global Average Pooling (GAP). CAM is not a re-ID task, but it's ability to locate objects lying within the feature map allows better representation learning in many vision tasks.
%\par Usually works adopting the human pose estimation approach require external annotations, while body part detection need no assistant from additional dataset.

%\subsection{Person attribute approach}
%Zhou \etal \cite{yin2018adversarial} used the adversarial technique to learn a semantically discriminatory joint space of a person image and generate an aligned analogous image for high level attribute image. Wang \etal \cite{wang2018transferable} proposed an unsupervised learning of transferring the the labelled information of a dataset to another unlabelled target domain for person re-id prediction. Both identity and attributes are learned and transferred. Wang \etal \cite{wang2017attribute} developed a Joint Recurrent Learning (JRL) model that is able to learn jointly attribute sequential ordering dependencies in the end-to-end encoder-decoder network.

\section{Proposed Attribute Attention Network (AANet)}
\label{sec:aanet}
%We first explained our AANet high level architecture, then we elaborate on each of the tasks. We conclude this section with loss calculation and implementation.

%\subsection{AANet Architecture Design}
The proposed AANet is a multitask network with three sub-networks, namely the GFN, PFN and AFN (Figure~\ref{fig:AANet}). The sub-network, GFN, performs global image-level ID classification. The PFN detects and extracts localized body parts before the classification task. The AFN uses person attributes for the classification task and generates the Attribute Activation Map (AAM) that plays a crucial role in identity classification. Some examples of AAM are shown in Figure \ref{fig:AAM_Heatmap}. In the figure, the generated AAMs provide more discriminatory features than the ID heatmap. As a result, when GFN, PFN and AFN learn together, our AANet becomes more generic and better at predicting person ID. The various components of AANet are described in details in the following sections.

\par The backbone network of AANet is based on ResNet architecture (Figure~\ref{fig:AANet}) since ResNet is known to perform well in re-ID problems. We removed the fully connected layer of the backbone network so that AANet's sub-networks can be integrated. There are four classifiers within AANet. They are the Global ID Classifier, Part Classifiers, Attribute Classifiers and AAM Classifier. The Global ID and Part classifiers belong to GFN and PFN respectively. The Attribute and AAM Classifiers belong to AFN. All four classifiers have rather similar network design. All of them utilize global average pooling to reduce over-fitting, and there is a 3 layers (Z, V and C) architecture to increase network depth for better feature learning. The classifiers learn using Softmax normalization and Cross-entropy loss.

\subsection{Global Feature Network}
This network performs the identity (ID) classification using the query image (Figure~\ref{fig:AANet}). The convolutional feature map $X \in R^{Z\times{H}\times{W}}$ extracted by the backbone network is provided as input to a global average pooling (GAP) layer. This is followed by a 1x1 convolution layer that brings the dimensionality down to V. BatchNorm and Relu are then applied to V before linear transformation to C, which is used by Softmax function. Cross-entropy loss is calculated on Softmax output for learning using back-propagation.

%\begin{figure}
%\centering
%\includegraphics[width=\linewidth]{images/GFN.jpg}
%\caption{GFN is the main task of AANet. It performs global ID classification. During the forward pass, the input feature map \(X\) goes through GAP, then a 1x1 convolution to reduce the channel depth from z to v, then to the fully connected layer with the dimension similar to the number of training identities. Best viewed in color}
%\label{fig:GFN}
%\end{figure}

\subsection{Part Feature Network}
\begin{figure}
\centering
\includegraphics[width=\linewidth]{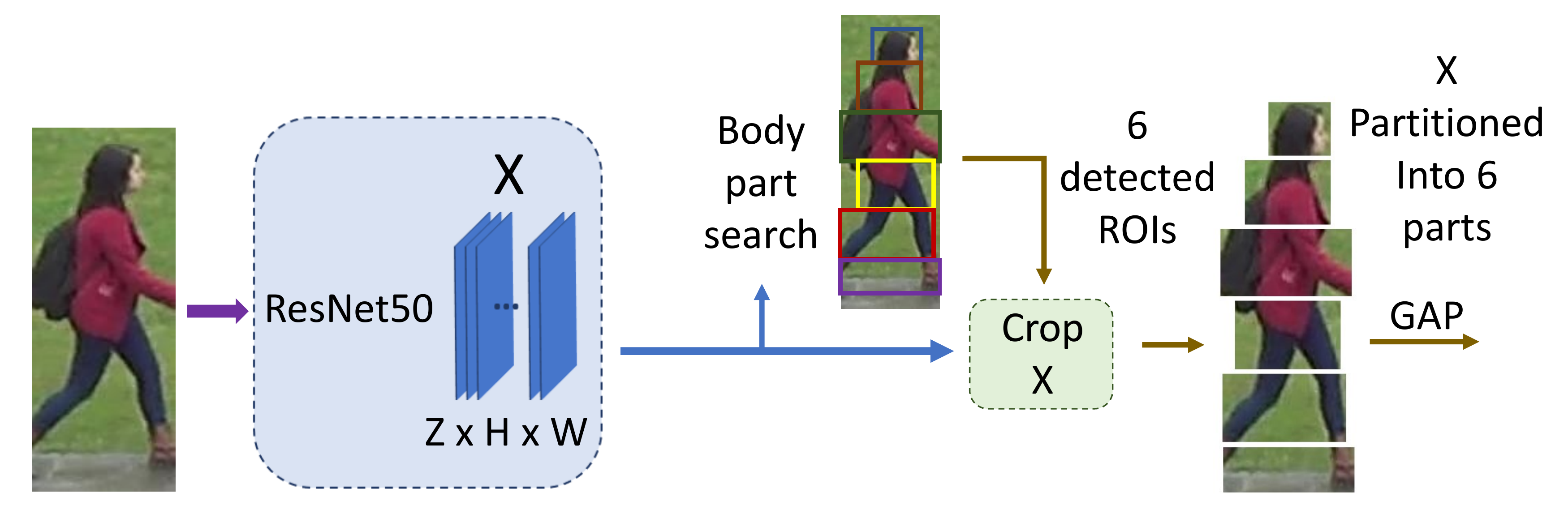}
\caption{The PFN divides the feature map $X\in R^{Z\times{H}\times{W}}$ into six ROIs using peak activation detection and pooling. Features from these $6$ ROIs are further used for identity classification. Best viewed in color.}
\label{fig:PFN}
\end{figure}

This network performs ID classification on body parts using the same person ID labels used in GFN. The architecture is shown in Figure \ref{fig:PFN}. The body part detector partitions the convolutional feature map X into six horizontal parts and estimate the corresponding regions of interest (ROIs). This is done by identifying peak activation regions in $X$. Let $(h_z, w_z)$ denote the peak activation location in each feature map $z$ of $X$ where $z\in\{1,\hdots ,Z\}$. 
\begin{equation}
(h_z,w_z) = \arg\max_{z}X_z(h,w)
\end{equation}
where $X_z(h,w)$ is the activation value at location $(h,w)$ on the $z$'th feature channel of $X$. These locations are then clustered into $6$ bins based on their vertical positions. These $6$ bins constitute the $6$ ROIs/ parts. The feature map $X$ is now divided into $6$ parts using these ROIs. Figure~\ref{fig:PFN} shows this process. Once the $6$ parts are computed, the subsequent processing, which is shown in Figure \ref{fig:AANet} is performed similarly as in GFN. 

\begin{figure}
\centering
\includegraphics[width=\linewidth]{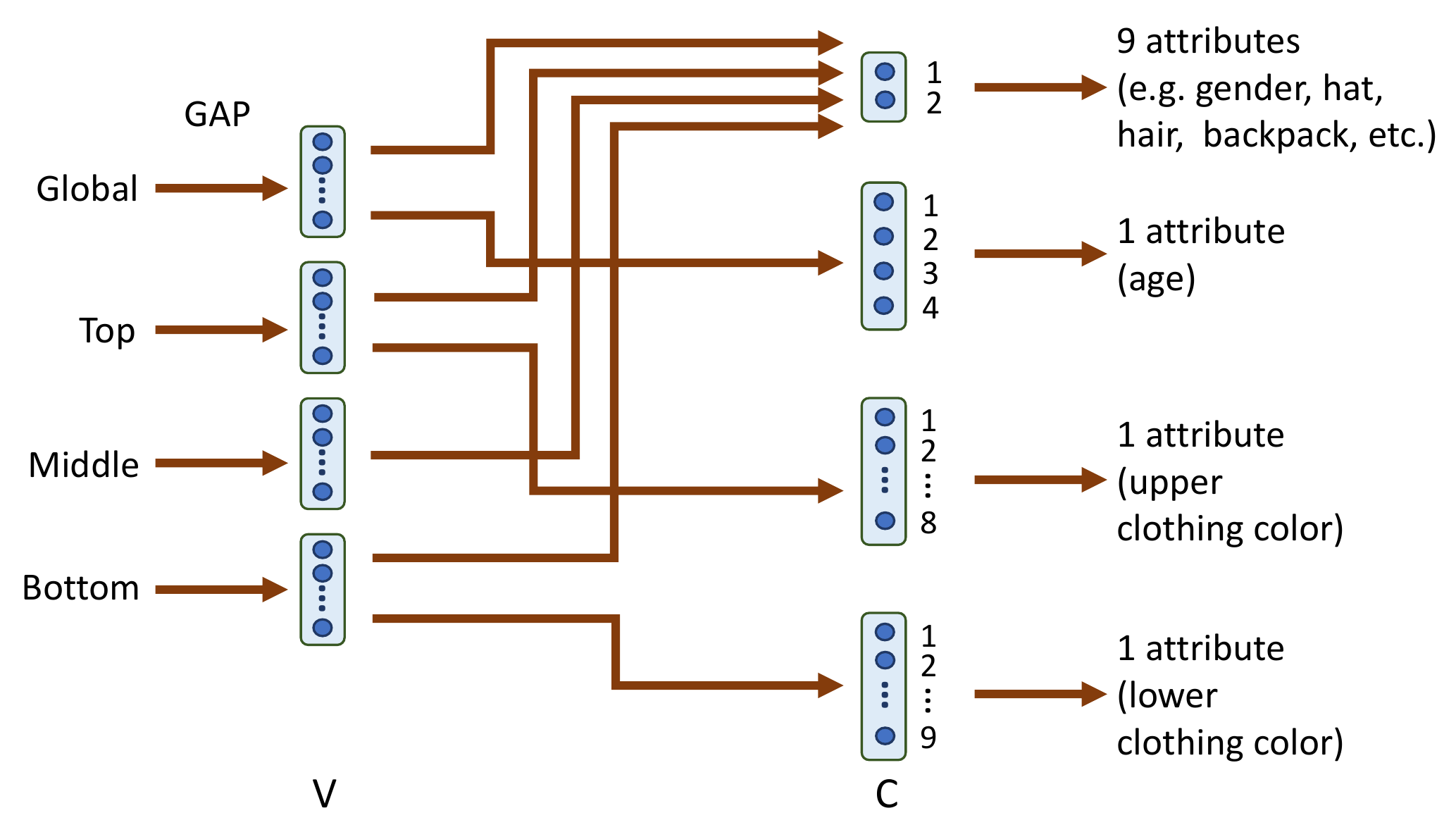}
\caption{12 attributes are generated from global, top, middle and bottom vectors on Market1501 dataset.}
\label{fig:attr_classifiers}
\end{figure}

\begin{figure*}
\centering
\includegraphics[width=\linewidth]{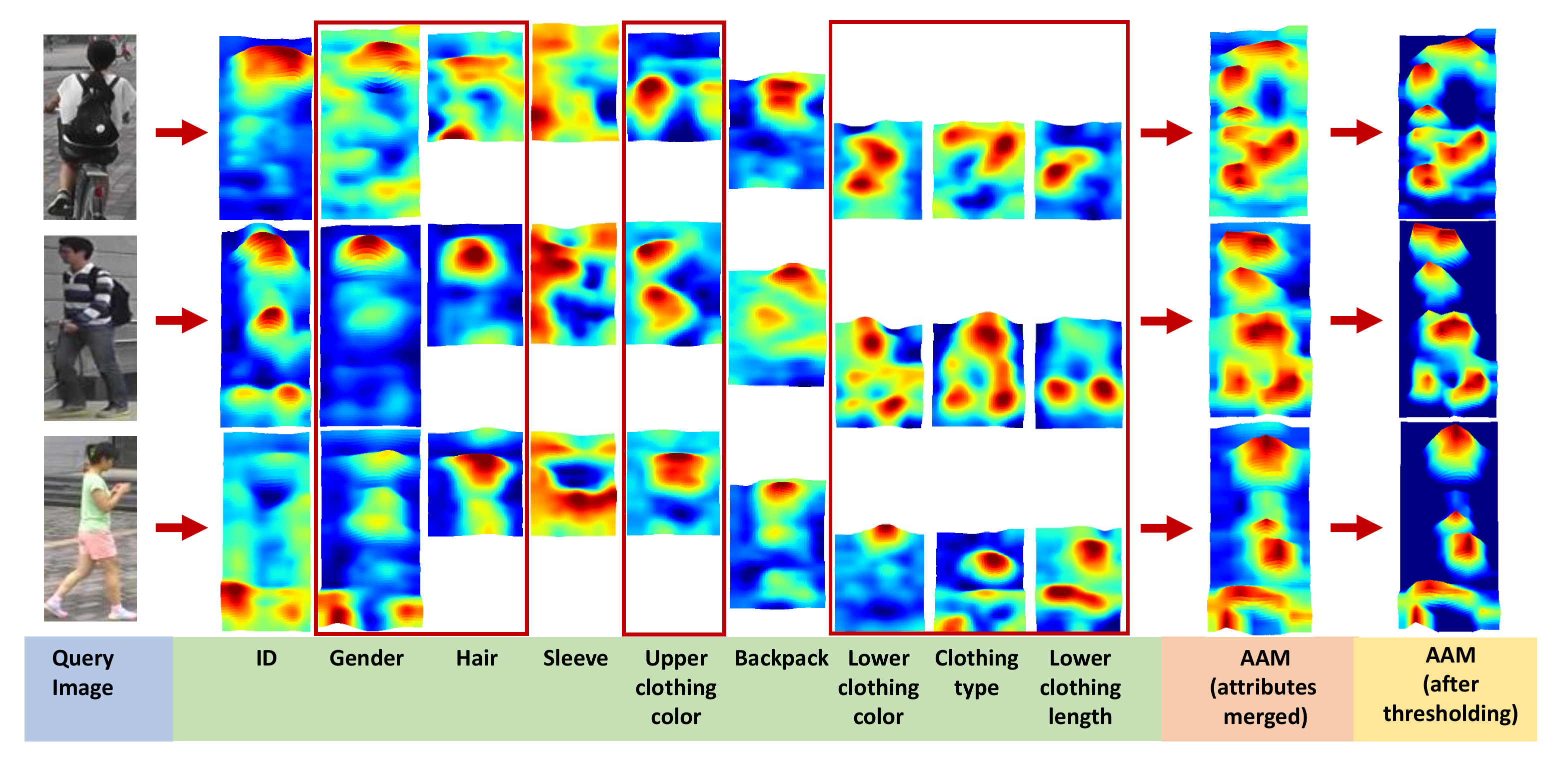}
\caption{On Market1501 dataset, a total of 12 attention maps are used. Eight most important attributes are shown. Only visual cues (in red boxes) are selected for AAM generation. Global attention map obtained from the GFN is shown here for qualitative comparison with other attributes. Backpack, handbag, bag and hat attributes, despite being important visual cues, do not appear in all images, and are therefore dropped. 
Attention map for sleeves captures too much background information and thus is unsuitable for AAM. 
Best viewed in color.}
\label{fig:attrs_to_aam}
\end{figure*}

\subsection{Attribute Feature Network}

The AFN captures the key attribute information in the AANet architecture (Figure~\ref{fig:AANet}). The AFN consists of two sub-tasks (i) attribute classification and (ii) attribute attention map (AAM) generation. The first sub-task performs classification on individual person attributes. The second sub-task leverages on the output of first sub-task and generates class activation map (CAM) \cite{zhou2016learning} for each attribute. CAM is a technique to localize the discriminatory image regions even though the network is trained on image-level labels only. Thus CAM fits well for AANet use. The CAMs generated from selected attribute classes are combined to form a feature map that is forwarded to the AAM Classifier for learning. We describe these two sub-tasks in the following paragraphs in detail. 

\textbf{(i) Attribute classification} 
The first sub-task of AFN is to perform attribute classification. There are 10 and 12 annotated attributes on DuketMTMC-reID and Market1501 respectively. The first layer of AFN is a 1x1 convolution that downsized the channel depth of feature map X from Z to V. Next, we partition the feature maps into three different sets, namely the Top, Middle and Bottom feature maps, each responsible for extracting features from their localized regions. Part-based modeling is known to reduce background clutter and improve classification accuracy. The different parts focus on different attributes. The Top feature maps, for example, are used for capturing features such as hat, hair, sleeves and upper clothing color etc. Features from the lower half of the body are ignored in the Top feature map. As shown in Figure \ref{fig:attr_classifiers}, the outputs of these feature maps, together with global feature map, are average pooled to generate 4 feature vectors at layer V. These 4 vectors are the input to the fully connected layer C. On Market1501, there are 4 classifiers at layer C, each generating their own attribute predictions. 

\textbf{(ii) Attribute Attention Map} The Attribute Attention Map (AAM) is the input to the Attribute Classifier (Figure \ref{fig:AANet}, which performs person ID classification. AAM combines class sensitive activation regions from individual attributes. These individual class-sensitive activation regions are extracted using CAM from each person attribute. As explained before, CAM uses GAP, with little tweak, to generate discriminatory image regions. Thus, CAM's output reveals image regions representing the attribute. Figure~\ref{fig:attrs_to_aam} shows some example of class sensitive activation regions and the combined AAM. For qualitative comparison, the second column in the figure shows the activation map generated by the global identification task (GFN). The subsequent columns show the class specific activation regions of various attributes such as gender, hair, sleeve, upper clothing color etc. The sixth column, for example, depicts the class specific activation region for upper clothing color. We can observe that the activation region corresponds to the upper clothing region in the input query image. 

\par Out of the 12 available attributes, the gender, hair, upper and lower clothing colors, lower clothing type and length are good choices for AAM generation. The AAM generation process involves merging the individual class specific activation regions by maximum operation and performing an adaptive thresholding. The thresholding process removes some background regions that sometimes appear within the class specific activation region. An example of this can be observed in the second row of Figure~\ref{fig:attrs_to_aam} where the Lower clothing activation map contains some background region but on thresholding the region is removed from the generated AAM. When qualitatively comparing the class activation map generated by the global feature network on the same query image, we can see that the AAM was more specific in localizing regions with distinct attribute information. The Attribute Classifier (Figure \ref{fig:AANet}) takes the AAM and perform ID classification, and shares the learning experience with the GFN and PFN.

\subsection{Loss calculation}
The proposed AANet is formulated as a multitask network. The multitask loss function of the AANet is defined as follows:
\begin{align}
%  L_{global}(y^G,d) & = %-\sum_{i=1}^{C}log(\frac{exp(y_i^G)}{\sum_{c=1}^{C}exp(y_c^G)}q(i)
    \mathcal{L}_{total}(x,W,\lambda) = \sum_{i=0}^{T}\lambda_i\mathcal{L}_i(x,W)
\end{align}
Where x is a set of training images, \(W\) is the weights on input \(x\). \(T\) is the total number of task loss \(\mathcal{L}_i\). \(\lambda_i\) are the task loss weighting factors,
%\(\mathcal{L}_{total}\) is thus given as
%\begin{align}
%    \mathcal{L}_{total} = \lambda_g\mathcal{L}_{global} + %\lambda_p\mathcal{L}_{part} +
%    \lambda_a\mathcal{L}_{attr} + \lambda_{aam}\mathcal{L}_{aam}
%\end{align}
%Where \(\mathcal{L}_{global}\), \(\mathcal{L}_{part}\), \(\mathcal{L}_{attr}\) and \(\mathcal{L}_{aam}\) are the task losses of GFN, PFN and AFN (attribute classification and AAM) respectively. All task losses are computed as cross-entropy losses during training. % The contribution of each loss term $(\lambda_g,\lambda_p,\lambda_{attr},\lambda_{aam})$ for optimal identity classification accuracy is automatically computed using homoscedastic uncertainty learning. 
and it plays an important role in optimizing the accuracy performance of AANet. If we assign equal weighting to all \(\lambda_i\), the retrieval accuracy will not be optimal. In our work, we used homoscedastic uncertainty learning \cite{journals/corr/KendallGC17, journals/corr/GalG15, kendall2017uncertainties} to obtain the task loss weighting.
We define the Bayesian probabilistic model classification likelihood output as 
\begin{align}
    p(y|f^W(x),\sigma) = Softmax(\frac{1}{\sigma^2} f^W(x))
\end{align}
Where \(f^W(x)\) is the output of the neural network and is scaled by \(\sigma^2\). \(\sigma\) is the observation noise. The log likelihood for this output is given by
\begin{equation}\label{equation_log_likelihood_1}
\begin{split}
    log(p(y=c|f^W(x),\sigma)) = \frac{1}{\sigma^2}f^W_c(x) \\
    - log(\sum_{i=0}^{C}exp(\frac{1}{\sigma^2}f_i^W(x)))
\end{split}
\end{equation}

Where \(C\) is the number of classes for the classification task. The task loss \(\mathcal{L}(x,W,\sigma)\) can be formulated as \(-log(p(y=c|f^W (x),\sigma))\). What we need is the cross-entropy loss of the non-scaled y, which if defined as \(\mathcal{L}(x,W)\) = \(-log\) Softmax(y,\:f\(^{W}\)(x)) \cite{journals/corr/KendallGC17}, the loss function can be simplified to 
\begin{align}
    \mathcal{L}(x,W,\sigma) \approx \frac{1}{\sigma^2}\mathcal{L}(x,W) + log \: \sigma
\end{align}

By applying the above loss function to our AANet, the final AANet loss function is now given as
\begin{equation}
\begin{split}
    \mathcal{L}(x,W,\sigma_g,\sigma_p,\sigma_a, \sigma_{aa}) \approx \frac{1}{\sigma^2_g}\mathcal{L}_g(x,W) + \frac{1}{\sigma^2_p}\mathcal{L}_p(x,W) +\\
    \frac{1}{\sigma^2_a}\mathcal{L}_a(x,W) + \frac{1}{\sigma^2_{aa}}\mathcal{L}_{aa}(x,W) +log\:\sigma_g\sigma_p\sigma_a\sigma_{aa}
\end{split}    
\end{equation}

Where \(\mathcal{L}_g\), \(\mathcal{L}_p\), \(\mathcal{L}_a\) and \(\mathcal{L}_{aa}\) represent global, part, attribute and attribute attention loss respectively. \(\sigma_g\), \(\sigma_p\), \(\sigma_a\) and \(\sigma_{aa}\) represent observation noises for global, part, attribute and attribute attention tasks respectively, and are inversely proportional to \(\lambda_i\)

\subsection{Implementation}

We implemented AANet with ResNet-50 and ResNet-152 as backbone networks, and pre-trained them with the ImageNet \cite{DBLP:conf/cvpr/DengDSLL009} dataset. Training images are enlarged to 384 x 128, with only random flip as the data augmentation method. Batch size is set to 32 for ResNet-50, and 24 for ResNet-152. Using Stochastic Gradient Descent (SGD) as the optimizer, we train the network for 40 epoch. Learning rate starts at 0.1 for the newly added layers, and 0.01 for the pretrained ResNet parameters, and follows the staircase schedule at 20 epoch with a 0.1 reduction factor for all parameters. In all the three sub-networks, the value of Z is 2048 and that of V is 256. The value of C depends on the dataset under evaluation. For DukeMTMC-reID \cite{ristani2016performance}, C is 702 and for Market1501 \cite{DBLP:conf/iccv/ZhengSTWWT15} it is 751. 

During testing, we concatenate the outputs of the V layer from all the classifiers, namely, the global identity classifier, the body part classifier, the attribute classifier and the AAM classifier (Figure~\ref{fig:AANet}) to form the representation of the query image. For ranking, we use the \(l_2\) norm between these descriptors of the query and the gallery images.

\section{Experimental Results}
\label{sec:results}
In the following experiments, we used the DukeMTMC-reID \cite{ristani2016performance} and Market1501 \cite{DBLP:conf/iccv/ZhengSTWWT15} datasets to conduct our training and testing. DukeMTMC-reID is a subset of DukeMTMC dataset. The images are cropped from videos taken from 8 cameras. The dataset consists of 16,522 training images and 17,661 gallery images, with 702 identities for both training and testing. 408 distractor IDs are also included in the dataset. There are a total of 23 attributes annotated by Lin \etal \cite{DBLP:journals/corr/LinZZWY17}. We use all attributes, but with modification to the clothing color attributes. We merged all 8 upper clothing color attributes and 7 lower clothing color attribute into a single upper clothing attribute and a single lower clothing attribute respectively. 

\par For Market1501, there are a total of 32,668 images for both training and testing. There are 751 identities allocated for training and 750 identities for testing. Lin \etal \cite{DBLP:journals/corr/LinZZWY17} also annotated this dataset, but with 27 person attributes. We use the same clothing color strategy as in DukeMTMC-reID, and use all attributes for training our model.

\subsection{Comparison with existing methods}

\par\textbf{DukeMTMC-reID dataset} We perform comparison with the state-of-the-art methods in Table~\ref{table:compare_duke}. The table has three parts based on the backbone network being used. First part compares models based on ResNet-50. The three comparative networks are KPM (Res-50) \cite{shen2018end}, PCB (Res-50)\cite{Sun_2018_ECCV} and the proposed AANet-50. We outperform the best state-of-the-art method \cite{Sun_2018_ECCV} in this category by 3.36\% in mAP and 3.12\% in Rank-1 accuracy.

\par The second comparison is based on networks using larger backbone models, which include both ResNet-101 and ResNet-152. The three comparative networks are GP-reID (Res-101) \cite{almazan2018re}, SPReID (Res-152) \cite{kalayeh2018human} and the proposed AANet-152. Here, we again outperformed the state-of-the-art method \cite{kalayeh2018human} by 0.95\% in mAP and 1.70\% in Rank-1 accuracy.

\par The third comparison is based on networks from the second comparison, but this time with re-ranking \cite{DBLP:conf/cvpr/ZhongZCL17}. We outperformed the state-of-the-art method \cite{almazan2018re} by 1.27\% in mAP and 0.96\% in Rank-1 accuracy.

\setlength{\tabcolsep}{8pt}
\begin{table}
\begin{center}
\begin{tabular}{l c c}
\hline\noalign{\smallskip}
\multicolumn{3}{c}{DukeMTMC-reID}\\
Methods & mAP & Rank-1\\
\noalign{\smallskip}
\hline
\noalign{\smallskip}
%LOMO+XQDA \cite{DBLP:conf/cvpr/LiaoHZL15} & 17.0 & 30.8\\
%GAN	\cite{DBLP:conf/iccv/ZhengZY17} & 47.1 & 67.7\\
%APR \cite{DBLP:journals/corr/LinZZWY17} & 51.9 & 70.7\\
%PAN	\cite{DBLP:journals/corr/ZhengZY17aa} & 51.5 & 71.6\\
FMN \cite{DBLP:journals/corr/abs-1711-07155} & 56.9 & 74.5\\
SVDNet \cite{DBLP:conf/iccv/SunZDW17} & 56.8 & 76.7\\
DPFL \cite{DBLP:conf/iccvw/ChenZG17} & 60.6 & 79.2\\
\hline
KPM (Res-50) \cite{shen2018end} & 63.2 & 80.3\\
PCB (Res-50)\cite{Sun_2018_ECCV} & 69.2 & 83.3\\
{\bf Proposed AANet-50}	& {\bf72.56} & {\bf86.42}\\
\hline
GP-reID (Res-101) \cite{almazan2018re}  & 72.80 & 85.20 \\
SPReID (Res-152) \cite{kalayeh2018human} & 73.34 & 85.95 \\ 
{\bf Proposed AANet-152}	& {\bf74.29} & {\bf87.65}\\
\hline
GP-reID (Res-101)\cite{almazan2018re} + RR & 85.60 & 89.40 \\
SPReID (Res-152)\cite{kalayeh2018human} + RR & 84.99 & 88.96 \\ 
{\bf Proposed AANet-152 + RR} & {\bf86.87} & {\bf90.36}\\
\hline
\end{tabular}
\end{center}
\caption{Performance comparison with other state-of-the-art methods using DukeMTMC-reID dataset. AANet-50 denotes AANet trained using ResNet-50. AANet-152 denotes AANet trained using ResNet-152. RR denotes Re-Ranking\cite{DBLP:conf/cvpr/ZhongZCL17}
\label{table:compare_duke}.}
\end{table}
\setlength{\tabcolsep}{1.4pt}

\par \textbf{Market1501 dataset} We perform similar comparisons as in previous section in Table~\ref{table:compare_market1501} using the Market1501 dataset. In the first comparison, which uses ResNet-50, the networks selected are KPM (Res-50) \cite{shen2018end}, PCB (Res-50)\cite{Sun_2018_ECCV} and the proposed AANet-50. We outperformed the best state-of-the-art method \cite{Sun_2018_ECCV} in this category by 0.85\% in mAP and 0.09\% in Rank-1 accuracy.

\par The second comparison is made using GP-reID (Res-101) \cite{almazan2018re}, SPReID (Res-152) \cite{kalayeh2018human} and the proposed AANet, with either ResNet-101 or ResNet152. Here, we again outperformed the state-of-the-art method \cite{kalayeh2018human} by 0.05\% in mAP and 0.25\% in Rank-1 accuracy.

\par The third comparison is made on networks from the previous section but with re-ranking~\cite{DBLP:conf/cvpr/ZhongZCL17}. We outperform the state-of-the-art method \cite{kalayeh2018human} by 1.42\% in mAP and 0.47\% in Rank-1 accuracy. We believe that the attribute information is a key contributor in AANet's person re-ID performance.

\setlength{\tabcolsep}{4pt}
\begin{table}
\begin{center}
\begin{tabular}{l C{1.0cm} C{1.1cm} C{1.1cm}}
\hline\noalign{\smallskip}
\multicolumn{4}{c}{Market1501}\\
Methods & mAP & Rank1 & Rank10\\
\noalign{\smallskip}
\hline
\noalign{\smallskip}
%PAR \cite{DBLP:conf/iccv/ZhaoLZW17} & 63.4 & 81.0 & 92.0 & 94.7\\
%MultiLoss \cite{DBLP:conf/ijcai/LiZG17} & 64.4 & 83.9 & -\\
PDC \cite{DBLP:conf/iccv/SuLZX0T17} & 63.4 & 84.4 & 94.9\\
PL-Net \cite{DBLP:journals/corr/YaoZZLT17} & 69.3 & 88.2 & -\\
DPFL \cite{DBLP:conf/iccvw/ChenZG17} & 73.1 & 88.9	& -\\
GLAD \cite{DBLP:conf/mm/WeiZY0T17} & 73.9 & 89.9 & -\\
\hline
KPM (Res-50)\cite{shen2018end} & 75.3 & 90.1 & 97.9\\
PCB (Res-50)\cite{Sun_2018_ECCV} & 81.6 & 93.8 & 98.5\\
{\bf Proposed AANet-50} & {\bf82.45} & {\bf93.89} & {\bf98.56}\\
\hline
GP-reID (Res-101) \cite{almazan2018re} & 81.20 & 92.20 & - \\
SPReID (Res-152) \cite{kalayeh2018human} & 83.36 & 93.68 & 98.40 \\ 
{\bf Proposed AANet-152} & {\bf83.41} & {\bf93.93} & {\bf98.53}\\
\hline
GP-reID (Res-101)\cite{almazan2018re}+RR & 90.00 & 93.00 & - \\
SPReID (Res-152)\cite{kalayeh2018human}+RR & 90.96 & 94.63 & 97.65 \\ 
{\bf Proposed AANet-152+RR} & {\bf92.38} & {\bf95.10} & {\bf97.94}\\
\hline
\end{tabular}
\end{center}
\caption{Performance comparison with other state-of-the-art methods using Market1501 dataset. AANet-50 denotes AANet trained using ResNet-50. AANet-152 denotes AANet trained using ResNet-152. RR denotes  Re-Ranking\cite{DBLP:conf/cvpr/ZhongZCL17}.}
\label{table:compare_market1501}
\end{table}
\setlength{\tabcolsep}{1.4pt}

\subsection{Network Analysis}
In this section, we study the effect of task loss weights and the size of the backbone network on the re-ID performance. We also review various training parameters.

\par \textbf{Ablation Study} In Table \ref{table:duke_ablation_study}, we show the impact of the task loss weights on AANet accuracy performance using DukeMTMC-reID dataset. The global ID task, the part task, the attribute classification task and the attribute attention map task are denoted as \(\mathcal{L}_g\), \(\mathcal{L}_p\), \(\mathcal{L}_a\) and \(\mathcal{L}_{aa}\) respectively. 
As we add each of these relevant tasks to the network, the accuracy improves, which justifies the contribution of each task on the overall performance. When we use homoscedastic uncertainty learning to obtain the task loss weights \(\mathcal{L}_g\), \(\mathcal{L}_p\) and \(\mathcal{L}_a\), the performance improves to 70.47\% mAP and 85.44\% Rank-1 accuracy. This result alone is enough to outperform the best state-of-the-art method using ResNet-50. With the integration of AAM, which provides more discriminatory features for learning, we improve the accuracy results to 72.56\% mAP and 86.42\% Rank-1 accuracy. 

\par \textbf{Effect of Backbone Network} 
%As shown by our AANet, Kalayeh \etal \cite{kalayeh2018human} and Almazan \etal \cite{almazan2018re}, 
The depth of the backbone network affects the accuracy performance of person re-ID. Deeper networks yield better result, and this is clearly shown in both Tables \ref{table:compare_duke} and \ref{table:compare_market1501}.  Table \ref{table:compare_duke} also shows that the our proposed smaller AANet-50 outperformed deeper SPReID (Res-152) \cite{kalayeh2018human} in Rank-1 accuracy by 0.47\%, and GP-reID (Res-101) \cite{almazan2018re} by 1.22\%. We achieved similar Rank-1 results on Market1501 dataset, with our AANet-50 outperforming those using deeper backbone networks. 

\par \textbf{Effect of Training Parameters} Many tricks have been used in the literature to enhance accuracy \cite{kalayeh2018human} and \cite{almazan2018re}. In \cite{kalayeh2018human}, the authors aggregate a total of 10 different datasets to generate $\sim$ 111k images and $\sim$ 17k identities for training and testing. In addition, multiple image sizes are used to train the network in different phases. In \cite{almazan2018re}, authors use techniques such as pre-training before regression learning, large image size, hard triplet mining and deeper backbone network for good person re-ID. These are good practices. However, the proposed AANet uses smaller image size, simpler training process, and a shallower ResNet-50 architecture to outperform existing state-of-the-art.
%mention about X dimension, number of parts
%Thresholding for both part and attribute

\setlength{\tabcolsep}{4pt}
\begin{table}
\begin{center}
\begin{tabular}{ | l | C{0.3cm} | C{0.3cm} | C{0.3cm} | C{0.3cm} | C{0.85cm} | C{0.85cm} |}
\hline%\noalign{\smallskip}
%\multicolumn{1}{l|}{} & \multicolumn{3}{c|}{} & %\multicolumn{2}{c|}{Classification} & \multicolumn{1}{c}{}\\
%\multicolumn{1}{|l|}{} & \multicolumn{3}{c|}{} & \multicolumn{2}{c|}{Accuracy}\\
%\cline{5-6}
\multicolumn{1}{|l|}{AANet-50} & \multicolumn{4}{c|}{Task} & \multicolumn{1}{c|}{mAP} & \multicolumn{1}{c|}{Rank 1}\\
\multicolumn{1}{|l|}{Task Loss} & \multicolumn{4}{c|}{Weights} & \multicolumn{1}{c|}{\%} & \multicolumn{1}{c|}{\%}\\
\hline
\(\mathcal{L}_g\) & 1 & 0 & 0 & 0 & 62.92 & 80.18\\
%\(\mathcal{L}_p\) & 0 & 1 & 0 & 74.3 & 91.0\\
%\(\mathcal{L}_a\) & 0 & 0 & 1 & - & -\\
%\(\mathcal{L}_g + \mathcal{L}_a\) & 1 & 0 & 1 & 75.5 & 90.3\\
\(\mathcal{L}_g + \mathcal{L}_p\) & 1 & 1 & 0 & 0 & 66.35 & 82.93\\
\(\mathcal{L}_g + \mathcal{L}_p + \mathcal{L}_a\) & 1 & 1 & 1 & 0 & 67.28 & 83.29\\
\hline
\multicolumn{1}{|l|}{\(\mathcal{L}_g + \mathcal{L}_p\ + \mathcal{L}_a\)} & \multicolumn{4}{c|}{Uncertainty} & \multicolumn{1}{c|}{70.47} & \multicolumn{1}{c|}{85.44}\\
\multicolumn{1}{|l|}{\(\mathcal{L}_g + \mathcal{L}_p + \mathcal{L}_a + \mathcal{L}_{aa}\)} & \multicolumn{4}{c|}{Learning} & \multicolumn{1}{c|}{\textbf{72.56}} & \multicolumn{1}{c|}{\textbf{86.42}}\\
\hline
\end{tabular}
\end{center}
\caption{Performance comparisons of different combination of task losses using DukeMTMC-reID dataset, with and without uncertainty learning. The top three rows are AANet accuracy with equal weights to the tasks. Bottom two rows show the results with loss weights obtained from uncertainty learning.
}
\label{table:duke_ablation_study}
\end{table}
\setlength{\tabcolsep}{1.4pt}

\begin{figure}
\centering
\includegraphics[width=\linewidth]{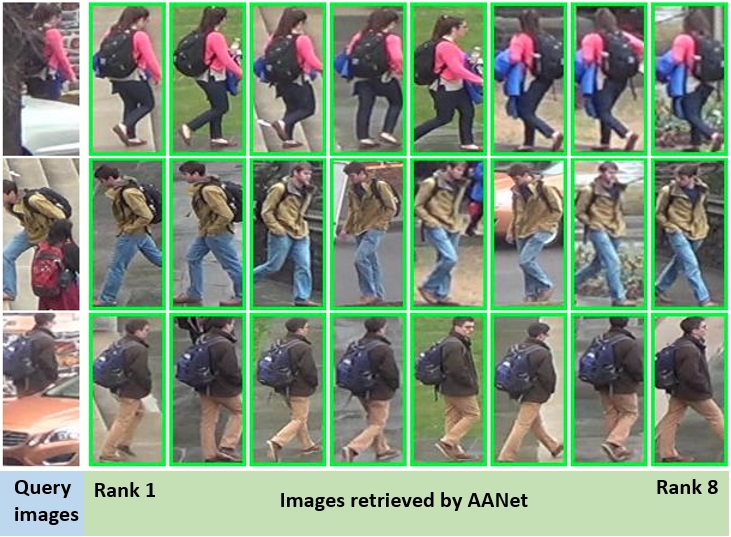}
\caption{Three queries from the DukeMTMC-reID with eight retrieved images for each query.}
\label{fig:good_retrieval}
\end{figure}

\begin{figure*}
\centering
\includegraphics[width=\linewidth]{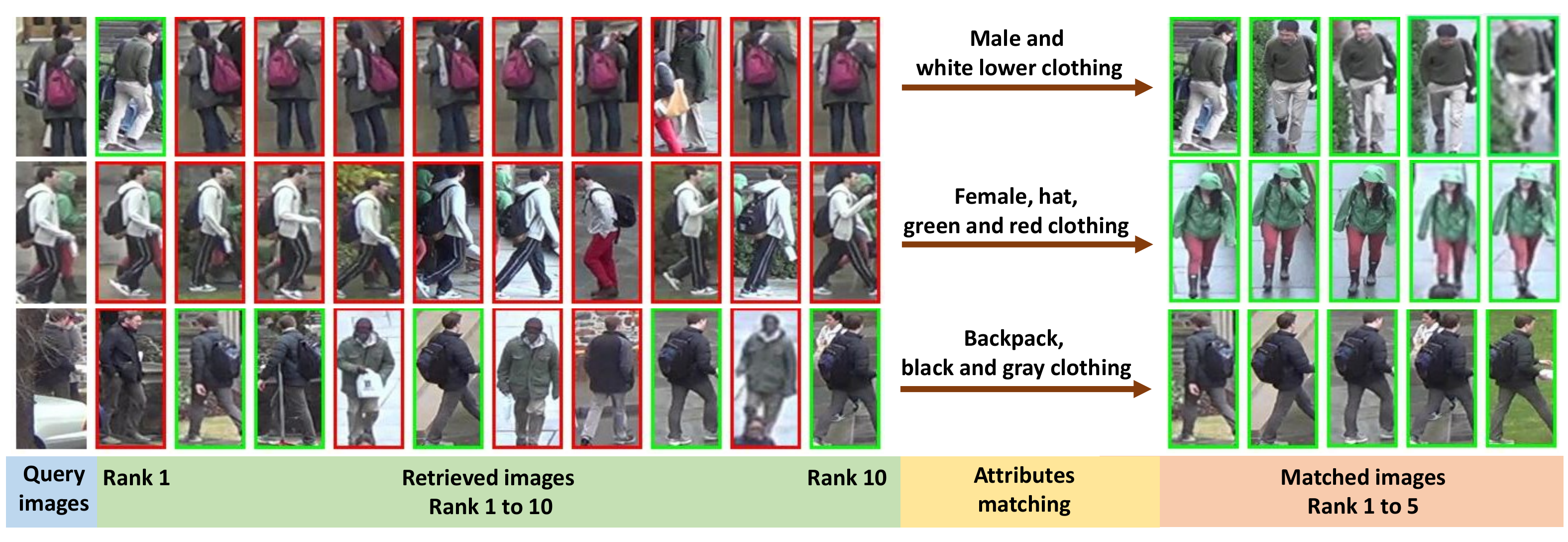}
%\includegraphics[scale=0.54]{images/attr_query.pdf}
%\caption{Three examples of how person attributes help in improving the image retrieval accuracy. First row used gender and color attributes to filter the retrieved images to get correct images within the top 5 ranks. Second row used hat, color and gender attributes to achieve same result as first row. Last row used color and backpack attributes to filter away falsely accepted images. Green box denotes same ID as query image. Red box denotes different ID from query image. Best viewed in color}
\caption{Three examples of how person attributes help in improving the image retrieval accuracy. These are challenging image queries that return many falsely accepted images. Since AANet returns each query and gallery images with predicted attributes, it provides an option for the user to use attribute matching to filter away the unwanted retrieved images. The useful attributes include gender. clothing colors, backpack, etc. Green box denotes same ID as query image. Red box denotes different ID from query image. Best viewed in color}
\label{fig:attr_query}
\end{figure*}

\setlength{\tabcolsep}{4pt}
\begin{table*}
\begin{center}
\begin{tabular}{ l | c c c c c c c c c c c c | c }
\hline
Methods & gender & age & hair & L.slv & L.low & S.clth & B.pack & H.bag & bag & hat & C.up & C.low & mean\\
\hline
APR \cite{DBLP:journals/corr/LinZZWY17} & 86.45 & 87.08 & 83.65 & 93.66 & 93.32 & 91.46 & 82.79 & 88.98 & 75.07 & 97.13 & 73.40 & 69.91 & 85.33\\
AANet-152 & {\bf92.31} & {\bf88.21} & {\bf86.58} & {\bf94.45} & {\bf94.24} & {\bf94.83} & {\bf87.77} & {\bf89.61} & {\bf79.72} & {\bf98.01} & {\bf77.08} & {\bf70.81} & {\bf87.80}\\
\hline
\end{tabular}
\end{center}
\caption{Performance comparisons of attribute accuracy on Market1501 dataset.
}
\label{table:attr_accuracy}
\end{table*}
\setlength{\tabcolsep}{1.4pt}

\section{Experimental Results Using Attribute}
\label{sec:results-attr}
In this section, we illustrate how person attributes help in refining the retrieved images for person re-ID. 

\subsection{Retrieval results}
We show three retrieval examples using AANet in Figure \ref{fig:good_retrieval}. Though there are some occlusions on the query subjects, for examples, cars and unwanted pedestrian, AANet has no problem retrieving correct images from the gallery set.
\par In Figure~\ref{fig:attr_query}, we show some examples of challenging queries where the subjects are heavily occluded. This resulted in poor retrieval accuracy. The figure demonstrates how AANet provides an option for the user to filter away the incorrect retrievals by using predicted attributes from query and gallery images. Three examples are given, each with their own retrieval difficulties. First example is given in row one. More than half of the query subject is occluded by another pedestrian. Most computer vision methods will pick the unwanted pedestrian as subject of interest, and return wrong images. In this example, 9 out of 10 images are wrongly retrieved, which results in poor mAP performance. Through AANet's attribute matching, those wrong images can be filtered out easily without laborious manual filtering. The ranking of theses attribute matched images were 1, 19, 38, 78, 172 during first retrieval, indicating how difference they are to query image. Same challenging queries are given in row two and three. As in first example, attribute filtering are performed to return correct images up to rank 5.
%image ID retrieval This is not just a tough re-ID task, which also a challenge to other computer vision domain like scene understanding. difficult to not just per, and the upper clothing color was similar to the other pedestrian clothing. Out of the top 10 images retrieved, only one correct image was retrieved. To filter away the wrong identities, we use person attribute query, by specifying male and white lower clothing color attributes as search criteria, and compare them with the predicted attributes of the retrieved images. AANet successfully returns correct identity images within the top 5 ranks. In the second row, the query subject is the lady in green and red clothing, but the retrieved images point to identities wearing white clothing only. The correct images are actually ranked from 14 to 21. To get the right images, we specified green and red clothing color attributes, gender attribute and hat attribute to effectively remove all those false retrievals, leaving us the images from correct identity. In last row, the subject is severely occluded, and has very similar attributes as found in those incorrect retrievals. We relied on black and gray clothing colors, and backpack attributes to retrieve all correct images within the top 5 rank.

\subsection{Attribute Classification Performance}
The accuracy of attribute classification of the proposed AANet is compared with APR \cite{DBLP:journals/corr/LinZZWY17} in Table \ref{table:attr_accuracy}. APR \cite{DBLP:journals/corr/LinZZWY17} is provided by Lin \etal, the author who annotated the DukeMTMC-reID and Market1501 datasets with person attributes. Since AANet employs localized attribute features to enhance network learning, 
%we are able to learn 
we obtained
better representations and outperforms APR in every attribute prediction. %It is worth noting that color related attributes, like the upper and lower clothing colors, do not perform well for both networks when compared to the rest of the attributes. The reason being cameras used to take videos or images are not color calibrated after installation. 
%Nonetheless, it is still very handy when performing image retrieval. 

%\setlength{\tabcolsep}{8pt}
%\begin{table}
%\begin{center}
%\begin{tabular}{l c c}
%\hline\noalign{\smallskip}
%\multicolumn{3}{c}{ResNet-50 AAM Design Parameters}\\
%Methods & mAP & Rank-1\\
%\noalign{\smallskip}
%\hline
%\noalign{\smallskip}
%\hline
%1 vector + no threshold  & 81.99 & 93.50 \\
%1 vector + threshold & 82.04 & 93.48 \\ 
%\hline
%4 vectors + no threshold  & 81.89 & 93.42 \\
%4 vectors + threshold & {\bf82.40} & {\bf93.87} \\
%\hline
%\end{tabular}
%\end{center}
%\caption{Performance comparison on AAM design parameters.}
%\label{table:aam_design_para}
%\end{table}
%\setlength{\tabcolsep}{1.4pt}

%\subsection{ResNet-50 AAM design parameters}
%We tested two crucial AAM parameters to validate our design assumptions. The first is the number of vectors used to generate the attributes. We tested single global vector vs the 4 global, top, middle and bottom vectors. We also tested the effectiveness of thresholding the AAM. The results are shown in Table \ref{table:aam_design_para}, which confirms that global and localized region vectors, and thresholding AAM give the best result. These parameters, in reality, reduce unwanted noise from our intended features, thus allowing our AANet to learn better.

\section{Conclusions}
\label{sec:conclu}
%We showed that our AANet, with the attribute attention map capturing the class sensitive regions, and attenuating the backgrounds, is able to share its unique learning experience with the global identity task, and compliment each others to bring out the very discriminatory features useful for person re-ID task. As a result, we outperformed the best state-of-of-art methods, regardless of the backbone architecture employed. On top of that, our AANet achieves good accuracy on person attribute prediction. This provides our user a very efficient and flexible image retrieval system.
%\par One area which we think we can pursue in the future is to leverage on those under utilized attributes to create a more complete person attention map.
%In this paper we propose a novel architecture to incorporate attributes based on physical appearance such as clothing color, hair, presence/absence of backpack etc. into a classification based person re-ID framework. The proposed Attribute Attention Network (AANet) employs joint end-to-end learning and uncertainty learning for multitask loss fusion. The resulting network outperforms existing state-of-the-art re-ID solutions on multiple benchmark datasets. In addition, AANet produces textual descriptions of the queried image based on the queried person’s physical attributes. This enables text-based query in addition to the traditional image-based query for re-ID. The AANet, to the best of our knowledge, is the first unified framework that allows both image and text-based queries for person re-ID.
In this paper we propose a novel architecture to incorporate attributes based on physical appearance such as clothing color, hair, %presence/absence of 
backpack etc. into a classiﬁcation based person re-ID framework. The proposed Attribute Attenion Network (AANet) employs joint end-to-end learning and homoscedastic uncertainty learning for multitask loss fusion. The resulting network outperforms existing state-of-the-art re-ID methods on multiple benchmark datasets.

\section*{Acknowledgement}This research was carried out at the Rapid-Rich Object Search (ROSE) Lab at the Nanyang Technological University, Singapore. The ROSE Lab is supported by the Infocomm Media Development Authority, Singapore.
{\small
\bibliographystyle{ieee_fullname}
\bibliography{egbib}
}
%\end{multicols}
\end{document}